\documentclass{ifacconf}

\usepackage{graphicx}      
\usepackage{natbib}        
\usepackage{xcolor}        
\usepackage{tabu}
\usepackage{tabularx}
\usepackage{adjustbox}
\usepackage{amsmath}
\usepackage{multirow}
\usepackage{wrapfig}
\usepackage{amsfonts}

\begin{document}
\begin{frontmatter}

\vspace{-0.2in}
\title{Integrating Reconfigurable Foot Design, Multi-modal Contact Sensing, and Terrain Classification for Bipedal Locomotion} 


\author[First]{Ted Tyler},
\author[First]{Vaibhav Malhotra}, 
\author[First]{Adam Montague},
\author[First]{Zhigen Zhao},
\author[First]{Frank L. Hammond III},
\author[First]{and Ye Zhao} 

\address[First]{George W. Woodruff School of Mechanical Engineering, Georgia Institute of Technology, 
   Atlanta, GA 30309 USA (e-mail: ttyler31, vaibhavmalhotra99, adam.montague, zhigen.zhao@gatech.edu and frank.hammond, ye.zhao@me.gatech.edu).}

\begin{abstract}                
The ability of bipedal robots to adapt to diverse and unstructured terrain conditions is crucial for their deployment in real-world environments. To this end, we present a novel, bio-inspired robot foot design with stabilizing tarsal segments and a multifarious sensor suite involving acoustic, capacitive, tactile, temperature, and acceleration sensors. A real-time signal processing and terrain classification system is developed and evaluated. The sensed terrain information is used to control actuated segments of the foot, leading to improved ground contact and stability. The proposed framework highlights the potential of the sensor-integrated adaptive foot for intelligent and adaptive locomotion. 
\end{abstract}


\end{frontmatter}

\section{Introduction}
\begin{figure}
\begin{center}
\includegraphics[width=8.4cm]{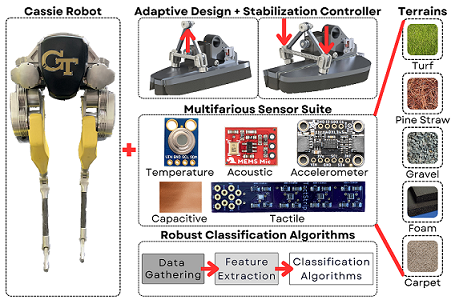}    
\vspace{-0.1in}
\caption{Illustration of a reconfigurable Cassie foot design with stabilizing tarsal segments and multi-modal sensing suite (i.e., multiple sensors for various features).}
\label{fig:overview}
\end{center}
\end{figure}

\vspace{-0.15in}
Legged robots are increasingly required to traverse a variety of environments and terrains to accomplish their tasks \cite{shamsah2023integrated}. For example, agricultural robots for fruit harvesting need to walk on grass, dirt, mud, snow, and mulch; disaster recovery robots could navigate through building rubble, ruined streets, and collapsed houses; and home-assistance robots may traverse rugs, carpets, and a variety of tile and hardwood floors. 
For each type of terrain, the robot foot should be robust enough to enable stable ground support. In addition, sufficient sensory input and processing capability are necessary to identify the terrains in order to adjust the control strategy of the robot.
This paper proposes a versatile foot design for the Cassie robot, integrated with a multi-modal sensor suite (see Fig.~\ref{fig:overview}), and employs machine learning algorithms to classify terrains based on the sensor data.


Terrain-adaptive foot design is essential for a bipedal robot's ability to traverse harsh terrains. The default foot of a Cassie robot is narrow and prone to sinking on deformable terrains. In contrast, a rigid wide foot design does not make full contact with the ground if the terrain is not flat. Therefore, we propose a nature-inspired foot design with tarsal segments that can be actuated according to the terrain types, which adapts to  terrain surfaces with soft and uneven properties.

The robust classification of the terrain relies on multiple sources of sensory input. The features extracted from the sensor data are input into a machine learning based terrain classification system. The signal acquisition and terrain classification framework is closely related to the previous work in~\cite{9115990}, but with a newly designed sensor suite involving temperature, acoustic, acceleration, capacitive, tactile measurements. Additionally, new feature extraction methods and deep learning algorithms are explored in this study.

The contributions for this paper lie in: (i) a nature-inspired foot design with adaptive foot stabilization over uneven terrains, complete with a high-level controller; (ii) the design of a tactile sensor with commercially available components, as well as the integration of the multifarious sensor suite and associated electronics; (iii) a terrain identification system capable of real-time operation on a Cassie robot, and an evaluation of the performances of various supervised learning algorithms for terrain classification.

\section{Related Work}
\vspace{-0.1in}
\textbf{Adaptive foot design: } 
Previous attempts have been made to design an adaptive foot with sensor integration to enhance robot locomotion. The works in~\cite{catalano2021adaptive} and~\cite{7803423} propose articulated soft foot designs that are compliant with unstructured terrains to maximize contact. The work in \cite{6290735} develops an actuated multi-contact foot with pressure sensor arrays integrated.

\textbf{Sensors for terrain classification: }
 A combination of inertial measurement units (IMUs), force, and torque sensors are commonly used to gather terrain data. Examples of this include~\cite{kolvenbach2019haptic}, \cite{wang2020situ}, \cite{bosworth2016robot}, \cite{venancio2020terrain}, \cite{walas2015terrain}, and \cite{walas2016terrain}. 
Visual data is also a popular input: \cite{walas2015terrain} combines force and torque data with visual and depth data, \cite{halatci2008study} uses a combination of images and wheel vibration in a rover, \cite{khan2011high} and \cite{khan2012visual} both use a single robot-mounted camera, and \cite{kurobe2021audio} uses camera and microphone input to identify terrain characteristics.
Tactile sensors, which mimic a human's ability to sense through touch, are also beneficial for terrain classification. ~\cite{9115990} proposes to use a variety of embedded sensors including tactile, acoustic, capacitive, temperature, and acceleration sensors, and shows that integrating multiple sensor modalities in general results in more effective terrain identification.


\textbf{Machine learning algorithms for terrain classification: }
Support vector machine is used by \cite{kolvenbach2019haptic}, \cite{walas2015terrain}, \cite{walas2016terrain}, \cite{halatci2008study}, and \cite{wu2016integrated}, while \cite{khan2011high}, \cite{khan2012visual}, and \cite{zhang2016terrain} all use the random forest algorithm. 
Recent trends indicate the potential of deep learning algorithms for terrain identification. Artificial neural networks (ANNs) are used in \cite{giguere2011simple}; Convolutional neural networks (CNNs) are used in \cite{venancio2020terrain}, \cite{kurobe2021audio}, and \cite{vulpi2021recurrent}; Recurrent neural networks (RNNs) are used in \cite{wang2020situ} and \cite{vulpi2021recurrent}; and transformers are used in \cite{bednarek2021fast} and \cite{bednarek2022haptr2}.

\section{Methods}
\vspace{-0.1in}
This section is organized in the following manner: the mechanical design and stability control of the adaptive foot is discussed in Sec.~\ref{mechanical}; the electrical design and sensor integration are discussed in Sec.~\ref{electrical}; the signal processing and terrain classification framework is discussed in Sec.~\ref{software}.

\begin{figure}[t]
\begin{center}
\includegraphics[width=8.4cm]{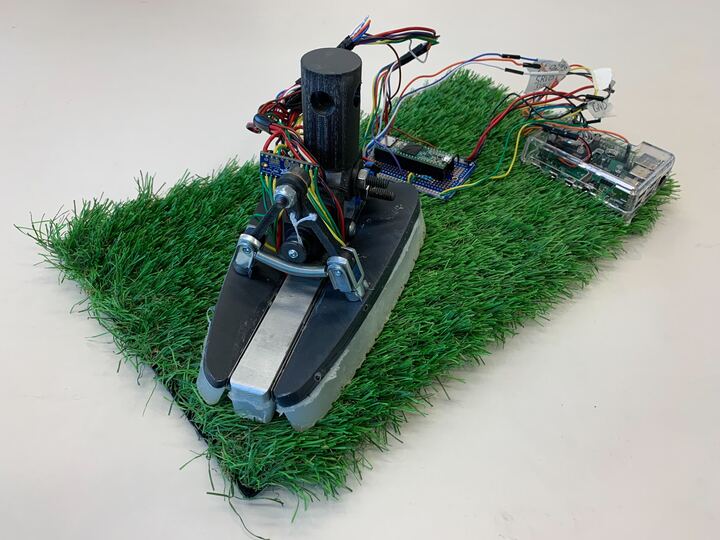}    
\caption{Prototype of a foot integrated with a Teensy 3.6 and Raspberry Pi.} 
\label{fig:prototype}
\end{center}
\end{figure}
\subsection{Mechanical Design}
\label{mechanical}
\vspace{-0.1in}
Two primary objectives are considered for the design of the foot. First, the foot must incorporate new tarsal segments that can adaptively stabilize the robot on rough terrain. Second, the foot needs to be lightweight to minimize deterioration of the Cassie robot's joints.



\begin{figure}[b]
\begin{center}
\includegraphics[width=8.4cm]{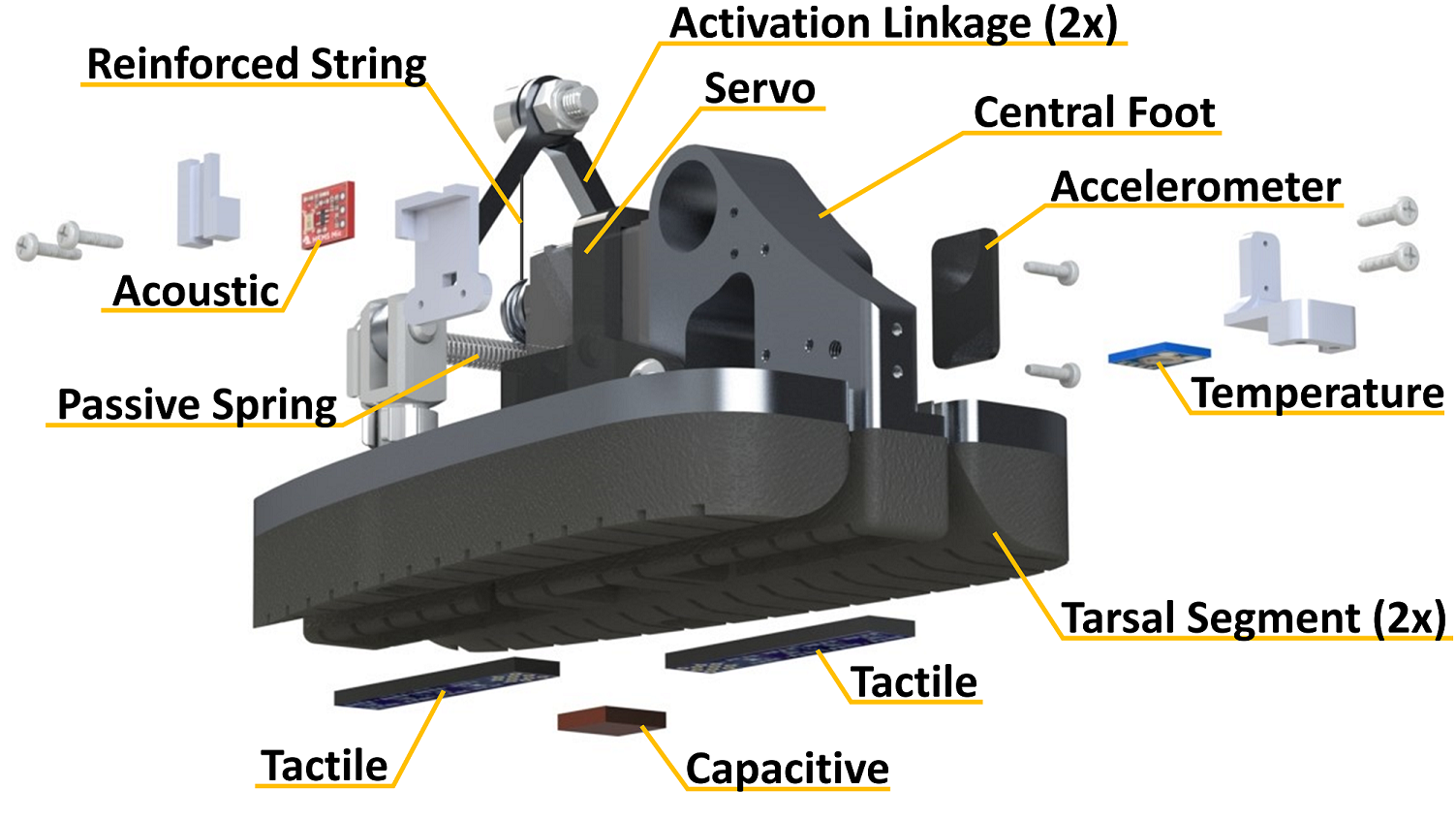}    
\caption{A labeled rendering of the foot that shows the placement of the sensors and mechanical components.} 
\label{fig:model}
\end{center}
\end{figure}

The design shown in Figs. \ref{fig:overview} and \ref{fig:model} fulfills the first objective by efficiently transferring rotational force from a servo into two tarsal segments on either side of the foot that act on the ground. The servo located at the center of the foot rotates to a selected angle between 0 and 360$^\circ$ based on the feedback from the sensors and the classification algorithm. As the servo rotates, the reinforced string attached to the servo's output shaft winds up, pulling the joint above the center of the foot down. This pulls on the linkages which directly push the tarsal segments down. The tarsal segments, which are attached to the central part of the foot using hinges, rotate to a maximum angle of fifteen degrees below the horizontal. When the foot is raised off the ground, the servo unwinds and the spring connecting the two tarsal segments passively retracts the tarsal segments to ten degrees above the horizontal. As such, the force from the servo acts through the linkages to pull directly on the tarsal segments. While it does have to overcome the force of the spring, it is a relatively weak spring that serves to barely lift the tarsal segments off the ground, preserving as much force from the servo for the tarsal segments as possible.

The servo provides a maximum torque of $22.8$ kg-cm, which is converted through the linkages in our design into a stabilizing force of up to $447$ N split between the two tarsal segments. If the robot is perfectly balanced and the tarsal segments are both flat on the ground, this force will be split evenly between the segments; as the robot falls to one side, the force will shift to the tarsal segment on that side to help the robot stay upright.


To actuate the tarsal segments, a bang-bang controller uses feedback from the classification algorithm, represented by the adaptive design portion of Fig. \ref{fig:overview}. The servo operates the tarsal segments if the algorithm predicts any of the following terrains that are potentially destabilizing: poppy seeds, gravel, pine straw, turf, a foam block, or carpet. As the tarsal segments descend, they increase the overall contact area of the foot by 292\% to distribute weight and stabilize the robot. As loose terrains shift, the tarsal segments can descend further to maintain contact with the ground. If a stable terrain is classified, the tarsal segment remains in its upright, passive position and only the central part of the foot is in contact.

To achieve the second goal of a lightweight foot, the physical components are fabricated using a variety of materials and methods. The tarsal segments, linkages, and actuator mount are all 3D printed using a nylon and carbon fiber filament, while the central component of the foot is milled using Aluminum 6061-T6. Other, less structurally essential components, like sensor mounts, are 3D printed using ABS filament. The final mass of all mechanical components of the foot is $0.64$ kg, while the total mass including electronics is $1.05$ kg. For reference, the original Cassie robot's foot is $0.21$ kg. A completed prototype of the foot is shown in Fig. \ref{fig:prototype}.

\subsection{Electrical Design}
\label{electrical}
\begin{figure}[b]
\begin{center}
\includegraphics[width=8.4cm]{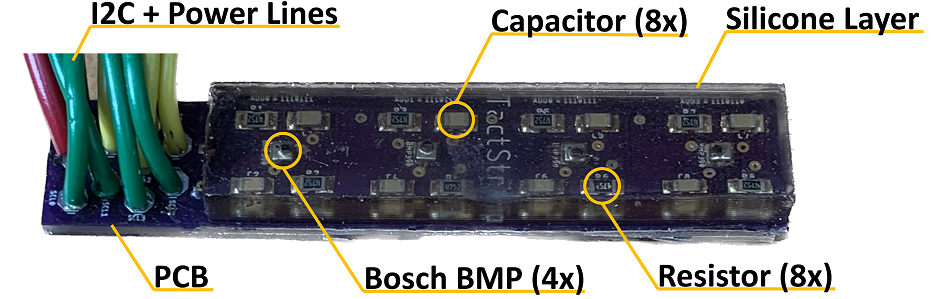}    
\caption{Closeup of a completed tactile sensor, with individual components labeled.} 
\label{fig:tactstrip}
\end{center}
\end{figure}

\vspace{-0.1in}
Five types of sensor are used on the foot: capacitive, acoustic, tactile, temperature, and accelerometer. These sensors and their placement around the foot are shown in Fig. \ref{fig:model}. The capacitive sensor is a plain copper plate connected to the microcontroller with a wire while the acoustic sensor (SPH8878LR5H-1, SparkFun), accelerometer (ADXL343, Adafruit), and temperature sensor (GY-906 MLX90614ESF, HiLetgo) are commercially available breakout boards. The new tactile sensors, like the one in Fig. \ref{fig:tactstrip}, are based on the design by \cite{tenzer2014feel} that uses barometric pressure sensors to register forces. We follow a similar fabrication process to the one outlined by \cite{koiva2020barometer} to construct the tactile sensors. However, two key differences from these papers are that we use Bosch BMP390 sensors (Bosch Sensortec), which communicate with the microcontroller through an I2C Multiplexer (TCA9548A, Adafruit) capable of coordinating with up to eight sensors with the same I2C address at once. To take full advantage of this multiplexer, two tactile sensors are used on the foot and each tactile sensor has four Bosch sensors.

A Raspberry Pi 4 microprocessor is used to provide sufficient processing power to perform data acquisition and terrain classification in real time. However, the Raspberry Pi 4 lacks analog to digital converters (ADC), making it incompatible with the requirements for the acoustic and capacitive sensors. Therefore, a Teensy 3.6 microcontroller is used alongside the Raspberry Pi. The Teensy gathers the data from the capacitive and acoustic sensors at 18 kHz and performs feature extraction. The Raspberry Pi reads and processes data from the other sensors at 45 Hz, and requests features extracted on the Teensy at the same frequency.


\subsection{Terrain Classification Algorithms}
\label{software}
\vspace{-0.1in}
\textbf{Preprocessing: }
The raw data is preprocessed and segmented so that the appropriate features can be extracted for use in classification. Before preprocessing, the raw tactile data experiences linear drift over time. To remove the drift and keep the input from each pressure sensor at a constant baseline, a detrending function is applied to the input of each sensor. This function performs a linear least-squares fit to the data and subtracts the solution from the data to remove the linear drift. The detrended tactile data is used to segment the rest of the data, i.e., split the data every time a footstep is taken so that statistical features can be extracted around the footsteps, and the unnecessary data between footsteps can be ignored. This is accomplished by identifying maxima and minima in the tactile data that corresponds to the foot in contact with the terrain.

\textbf{Feature Extraction: }
Feature extraction is applied to the segmented data so that each footstep is represented by a single set of variables for the machine learning algorithms. Using extracted features instead of raw data reduces the dimensionality of the data and facilitates the classification process. In total, 100 features are extracted from the five sensors using the equations below, where $\mathbf{x}=\{x_0, x_1, \dots, x_n\}$, and $x_{i} \in \mathbb{R}$ denotes the $i^{\rm th}$ scalar in a time series of sensor data with length $n$, and ZCR stands for the Zero-crossing Rate: 
\begin{align*}
     &\text{Sum: } f_{\rm sum}(\mathbf{x}) = \sum^{n}_{i=0} x_{i} \\
     &\text{Maximum: } f_{\rm max}(\mathbf{x}) = {\rm max}(x_1, x_2, x_3, ..., x_n) \\
     &\text{Minimum: } f_{\rm min}(\mathbf{x}) = {\rm min}(x_1, x_2, x_3, ..., x_n)   \\
     &\text{Mean: } \mu(\mathbf{x}) = \frac{1}{n} \sum^{n}_{i=0} x_i    \\
     &\text{Variance: } \sigma^2(\mathbf{x}) =  \frac{1}{n} \sum^{n}_{i=0} (x_i - \mu)^2  \\ 
     &\text{Skewness: } \tilde{\mu}^3 (\mathbf{x}) = \frac{1}{(n-1)^3} \sum^{n}_{i=0} (x_i - \mu)^3 \\
     &\text{Kurtosis: } f_{\rm kurt} (\mathbf{x})= \frac{1}{n(\sigma^2)^2} \sum^{n}_{i=0} (x_i - \mu)^4    \\
     &\text{Sign: } f_{\rm sign}(x_i) = \left\{
                                    \begin{array}{ll}
                                          0 \ x_i < 0 \\
                                          1 \ x_i\geq 0 \\
                                    \end{array} 
                                \right.   \\
     &\text{ZCR: } f_{\rm zcr}(\mathbf{x}) = \frac{1}{n} \sum^{n}_{i=1}|f_{\rm sign}(x_i) - f_{\rm sign}(x_{i-1})|   \\
     &\text{80\% Rise Time: } f_{80\%}(\mathbf{x}) = i \text{ for } x_i = 0.8(f_{\rm max}(\mathbf{x}))   \\
     &\text{Fast Fourier Transform: } f_{\rm fft}(\mathbf{x}) = FFT(\mathbf{x})
\end{align*}

The features extracted from each sensor are as follows:
\textit{Accelerometer: }the maximum, minimum, mean, variance, sum, and ZCR are calculated for each axis (X, Y, and Z) on the three-axis accelerometer for eighteen features.

\textit{Acoustic: }the ZCR and Fast Fourier Transform (FFT) functions are calculated. The FFT output is averaged over nine spectral bands for ten features.

\textit{Capacitive: }the mean and variance are calculated, as well as the same nine spectral band averages that are calculated for the acoustic sensor for 11 features.

\textit{Tactile: } the maximum value, minimum value, and 80\% rise time values are calculated for the data summed at each time step across the eight barometric pressure sensors. We also calculate the maximum, minimum, mean, variance, skewness, and kurtosis values for each individual sensor, as well as each individual sensor's value when the summed value reaches 80\% rise time for 59 features. As \cite{9115990} states, the summed value at 80\% of the rise time is used because it could be a valuable indication of the terrain's stiffness.

\textit{Temperature: } the mean and variance of the terrain's temperature during every footstep are extracted for two features.

\textbf{Classification Algorithms:}
We test four traditional machine learning algorithms and two deep learning algorithms to classify the extracted terrain features: k-nearest neighbor (KNN), support vector machine (SVM), random forest (RF), gradient boosting classifier (GB), artificial neural networks (ANN), and convolutional neural networks (CNN). Each classifier accepts the previously discussed extracted features as inputs for training. Techniques like K-Fold Cross Validation and Regression are implemented, as appropriate based on the algorithm, to find the best models.
\section{Experimental Setup}
\label{experiments}
\vspace{-0.1in}
Finite Element Analyses (FEA) are performed on the servo linkages, the central part of the foot, and the servo bracket to validate the mechanical design of the foot. To simulate the forces acting on the foot, the bottom of the foot is fixed and a force is applied at the ankle joint equal to the weight of the robot. The servo bracket is fixed by the holes where the bracket is bolted to the central foot, and the force required to secure the servo at the servo's maximum torque is applied to the upper, inside surface of the servo bracket. For the linkages, hinge constraints are applied to the inner cylindrical surfaces where they rotate as the servo actuates, and additional constraints are applied to represent the bolt fixing the linkages at the linkage joint above the servo. The force acting on the linkage is applied to the inside face of the upper hole parallel to the linkage pointing towards the opposite hole.  

\begin{table}[t]
\begin{center}
\caption{Hyperparameter Tuning}\label{tab:params}
\begin{tabular}{ l l c } 
\textbf{Algorithm} & \multicolumn{2}{ c }{\textbf{Hyperparameters}} \\
\hline
\multirow{4}{4em}{KNN} & \# of Neighbors & 10 \\ 
& Leaf Size & 30 \\ 
& Weight Function & Uniform \\ 
& Distance Computation & Minkowski \\
\hline
\multirow{3}{4em}{SVM} & Regularization Scaler & 1.0 \\ 
& Kernel & RBF \\ 
& Decision Function & One-VS-Rest \\
\hline
\multirow{3}{4em}{RF} & \# of Trees & 100 \\ 
& Maximum Depth & None \\ 
& Split Criteria & Gini Impurity \\
\hline
\multirow{4}{4em}{GB} & \# Loss Function & Logistic Loss \\ 
& Learning Rate & 0.1 \\ 
& \# of Estimators & 100 \\
& Max Depth & 3 \\
\hline
\multirow{6}{4em}{ANN} & Hidden Layers & 2 \\ 
& Nodes/Hidden Layer & 50, 100 \\ 
& Batch Size & 32 \\
& Activation Function & ReLU \\
& Optimizer & Adam \\
& Learning Rate & Adaptive \\
\hline
\multirow{7}{4em}{CNN} & Convolution Layers & 2 \\ 
& Batch-Norm Layers & 2 \\ 
& Max Pooling Layers & 2 \\
& Fully Connected Layers & 2 \\
& Trainable Parameters & 97,354 \\
& Batch Size & 32 \\
& Optimizer & Adam \\
& Learning Rate & $1\times 10^{-3}$ \\
\hline
\end{tabular}
\end{center}
\end{table}

\begin{wrapfigure}{l}{0.2\textwidth}
  \begin{center}
    \includegraphics[width=0.2\textwidth]{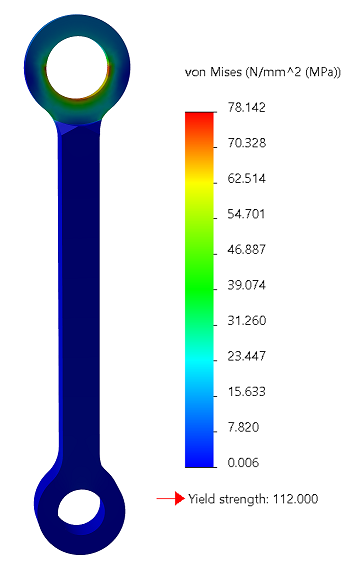}
  \end{center}
  \caption{FEA results for the linkage.}
  \label{fig:linkage}
\end{wrapfigure}

Fig. \ref{fig:linkage} shows the result of the FEA on the linkage as an example. Darker blue areas represent areas under lower stress, while red areas represent those with high stress. Because FEA results show that the linkages have the most potential for failure, an endurance limit study, which is related to fatigue failure, is performed for the linkages using Marin Factors. The results of these analyses are in Section \ref{results}. 

To gather training data for classification, we fabricate a prototype (Fig. \ref{fig:prototype}) and install it on an Instron 5965 machine, which can be programmed to simulate a footstep. For each step, the machine lowers the foot onto the terrain until $300$ N of force is measured. The foot is held in this position for one second while the tarsal segments, which actuate based on inputs from the tactile sensors, lowered. The machine then returns the foot to a neutral position as the tarsal segments retracted, where it rests for $0.25$ seconds before the process repeated. To avoid settling, we regularly shuffle any loose materials that shifts during data gathering to prevent erroneous readings. This cycle is repeated $1000$ times for each terrain so that each terrain is represented by $1000$ data points during classification. 

Ten materials are used in the experiments to simulate real-world terrains: a metal plate (METAL), a wood board (WOOD), a foam block (FOAM), a yoga mat (MAT), a patch of artificial grass (GRASS), gravel (GRAVEL), pine straw (STRAW), concrete (CONCRETE), a patch of carpet (CARPET), and poppy seeds (POPPY). We select these materials because they roughly represent a range of terrains that a bipedal robot could encounter in indoor and outdoor deployments. 

Once the training data is gathered, it is used to construct models for each of the algorithms discussed in Section \ref{software}. The 10,000 data points are divided and 80\% is allocated to a training set and 20\% is allocated to a test set. Each model is trained and tested on these same sets. Once these models are tuned and trained, they are saved and loaded onto the Raspberry Pi. To validate the feasibility of using our design in real-time applications, ten steps are timed and classified for each of the top three performing algorithms and the average classification speed is recorded. Table \ref{tab:params} outlines the set of hyperparameters that returns the highest test results for each model.

\vspace{0.3in}
\section{Results and Discussion}
\label{results}
\vspace{-0.1in}
\textbf{FEA Analysis: }
FEA results for the central foot, linkages, and servo bracket can be seen in Table \ref{tab:FEA}, which shows that the linkages have the most potential for failure with a factor of safety (FOS) of only 1.43. Despite the low FOS, the endurance limit study showed that the linkage design can withstand $5$x$10^8$ footsteps before failing. At the speed that the Cassie robot walks, this comes out to constant walking for 24 hours per day for over six years. As such, the strength of these elements is validated to be reliable.

\begin{figure}
\begin{center}
\includegraphics[width=6cm]{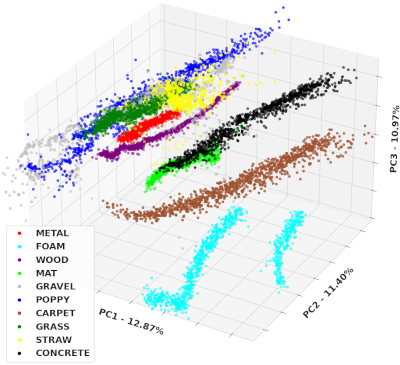}    
\vspace{-0.1in}
\caption{Visualization of features projected to 3D space using PCA. Each colored cluster represents the 100 features of each terrain projected onto three dimensions for visualization.} 
\label{fig:pca}
\end{center}
\end{figure}

\textbf{Learning Algorithm Evaluation: }
As shown in Table \ref{tb:offline_accuracy}, RF has the highest test accuracy with 99.90\%, while GB scores the second highest with 99.85\% accuracy. Both deep learning-based models score nearly as well with CNN scoring 99.85\% and ANN scoring 99.80\%. This high performance could be due to the significant distinction in features generated per class. We complete a principal component analysis (PCA) to reduce the feature space to three dimensions for visualization purposes. The resulting feature space, which can be seen in Fig. \ref{fig:pca}, clearly shows each class in relatively dense clusters, excluding a small number of outliers. This visualization illustrates how the feature extraction discussed in Section \ref{software} statistically groups the data from each terrain into unique clusters, making it simpler for the classification algorithms to identify the differences between each terrain. Based on the information obtained through the PCA, the high performance of both RF and GB also becomes clearer since tree-based models like RF and GB are known to outperform neural networks on tabular data, as seen in work by \cite{grinsztajn2022tree} and \cite{shwartz2022tabular}. SVM returned the lowest offline accuracy at 59.15\%. However, based on the PCA visualization, this also may not be surprising considering that SVM works by linearly separating one class from the rest, and generally works better with two or three classes at the most.

\begin{table}[b]
\centering
\caption{FEA Results}\label{tab:FEA}
\vspace{-0.1in}
\begin{tabular}{ m{1.8cm} m{1.7cm} m{1.8cm} m{1.3cm}  } \\
\textbf{Component} & \textbf{Max Stress (MPa)} & \textbf{Yield Stress (MPa)} & \textbf{Factor of Safety} \\
         \hline \\
         Central Foot & 10.87 & 275.00 & 25.3 \\
         Linkages & 78.14 & 112.00 & 1.43 \\
         Servo Bracket & 57.61 & 112.00 & 1.94 \\
         \hline
\end{tabular}
\end{table}

\textbf{Sensor Ablation Study: }
To understand how the classifiers learn, combinations of sensors are tested on RF and GB to see which sensors contribute the most to the classifiers. As Table \ref{tb:sensor_accuracy} shows, each sensor had similar effects on RF and GB accuracy. However, the accelerometer and tactile sensor data influence the classification algorithms more than the data of the other sensors. While this shows the robustness of the classification algorithms in being able to obtain sufficient information from only two sensors, it would be ideal if all of the sensors played a stronger role in the terrain identification to make the foot more resilient to failed or malfunctioning sensors. The reason for the disparity between sensors varies from sensor to sensor and could theoretically be caused by a number of reasons. More variation in the terrains, such as using hotter/colder and wet terrains, could increase the usefulness of the capacitive and temperature sensors. Moving the acoustic sensor away from the noisy servo could increase sound quality and make it more beneficial. And, as Section \ref{software} outlines, 59 of the 100 features extracted from the raw data come from the tactile sensor. The classifiers might simply rely more on tactile information because they received more statistical data from that sensor. As such, expanding the feature extraction of the other sensors could increase their usefulness to the classifiers.
\begin{table}[ht]
\begin{center}
\caption{Performance per Sensor}\label{tb:sensor_accuracy}
\vspace{-0.07in}
\begin{tabular*}{\columnwidth}{c c c}
\textbf{Sensor} & \textbf{RF Accuracy (\%)} & \textbf{GB Accuracy (\%)} \\
         \hline \\
         all & 99.90 & 99.85 \\
         temperature & 49.45 & 54.73 \\
         accelerometer  & 96.17 & 96.07 \\
         microphone  & 60.10 & 61.09 \\
         capacitive  & 44.08 & 44.63 \\
         tactile & 99.80 & 99.80 \\
         temp + mic + cap  & 86.22 & 85.67 \\
         temp + cap  & 69.40 & 71.19 \\
         temp + mic  & 82.84 & 83.08 \\
         mic + cap  & 68.26 & 68.71 \\
         \hline
\end{tabular*}
\end{center}
\end{table}
\vspace{-0.1in}

\begin{table*}[t]
\begin{center}
\caption{Test Accuracy Results (\%)}\label{tb:offline_accuracy}
\vspace{-0.1in}
\begin{tabular*}{\linewidth}{c c c c c c c c c c c c}
   & \textbf{Metal} & \textbf{Foam} & \textbf{Wood} & \textbf{Mat} & \textbf{Gravel} & \textbf{Poppy} & \textbf{Carpet} & \textbf{Grass} & \textbf{Straw} & \textbf{Concrete} & \textbf{Average} \\
         \hline \\
         RF & 100.00 & 100.00 & 100.00 & 100.00 & 99.49 & 100.00 & 100.00 & 99.50 & 100.00 & 100.00 & 99.90 \\
         GB & 100.00 & 100.00 & 100.00 & 100.00 & 99.49 & 100.00 & 99.53 & 99.50 & 100.00 & 100.00 & 99.85 \\
         CNN & 100.00 & 100.00 & 100.00 & 100.00 & 98.98 & 100.00 & 100.00 & 99.50 & 100.00 & 100.00 & 99.85 \\
         ANN & 100.00 & 100.00 & 100.00 & 100.00 & 100.00 & 99.52 & 100.00 & 99.50 & 99.04 & 100.00 & 99.80 \\
         k-NN & 100.00 & 100.00 & 100.00 & 100.00 & 64.97 & 52.86 & 99.53 & 81.50 & 46.89 & 100.00 & 83.96 \\
         SVM & 0.00 & 100.00 & 93.26 & 100.00 & 1.02 & 98.10 & 99.53 & 0.00 & 0.00 & 95.45 & 58.97 \\
         \hline
\end{tabular*}
\vspace{-0.1in}
\end{center}
\end{table*}

\textbf{Online Classification Speed: }
Models for GB, RF, and ANN are saved and loaded onto the Raspberry Pi to be timed. Although CNN outscores ANN by a slim margin, the accuracies are close enough that ANN's smaller model would run more efficiently on the microprocessor since CNN models generally are more computationally expensive. Over ten steps, GB, RF, and ANN have average classification speeds of 2.73 ms, 32.76 ms, and 0.64 ms, respectively. RF has the longest inference time, possibly due to a large number of trees (see Table \ref{tab:params}) within the model. ANN is the fastest by a large margin due to the relatively small size of the ANN model. Because classification is performed at the end of the footstep once the data for the step has been collected and processed, all of these times are fast enough that the data-gathering process of the following step would not be negatively impacted in a real-time application. This implies that the current system can realistically be used in an online terrain identification application.

\section{Conclusions}
\vspace{-0.1in}
Using multi-modal foot contact sensing consisting of an acoustic sensor, accelerometer, capacitive sensor, tactile sensor, and temperature sensor, the stabilizing Cassie foot was able to classify ten terrains with an average test accuracy of 99.90\%, demonstrating great potential for real-world applications. To achieve the objective of using this foot in these scenarios, the next step will be to assemble the foot onto the Cassie robot to obtain additional training data and complete real-time, online tests with the robot. Once the foot has successfully been integrated into the robot and tested, the classification algorithm will be used to update and optimize the robot's gait at runtime as it traverses a variety of terrains.


\bibliography{bibliography}             

\end{document}